\documentclass[10pt,twocolumn,letterpaper]{article}

\usepackage{iccv}
\usepackage{times}
\usepackage{epsfig}
\usepackage{graphicx}
\usepackage{amsmath}
\usepackage{amssymb}
\usepackage{lipsum}
\usepackage{subcaption}
\usepackage{booktabs}
\usepackage{colortbl}
\usepackage[export]{adjustbox}
\usepackage{multirow}
\usepackage[utf8]{inputenc}

\usepackage[pagebackref=true,breaklinks=true,letterpaper=true,colorlinks,bookmarks=false]{hyperref}
\usepackage{color}
\iccvfinalcopy 

\definecolor{cwblue1}{rgb}{0.27,0.427,0.623}
\definecolor{cwblue2}{rgb}{0.286,0.454,0.658}
\definecolor{cwblue3}{rgb}{0.733,0.811,0.905}

\newcommand{\bh}{\mbox{\boldmath$h$}}
\newcommand{\bm}{\mbox{\boldmath$m$}}

\newcommand{\bp}{\mbox{\boldmath$p$}}

\newcommand{\bv}{\mbox{\boldmath$v$}}

\newcommand{\bx}{\mbox{\boldmath$x$}}

\newcommand{\bH}{\mbox{\boldmath$H$}}
\newcommand{\bI}{\mbox{\boldmath$I$}}

\newcommand{\bM}{\mbox{\boldmath$M$}}

\newcommand{\bV}{\mbox{\boldmath$V$}}

\newcommand{\myparagraph}[1]{\textbf{#1} --- }  

\ificcvfinal\fi
\begin{document}
	
	\title{Human Action Recognition:\\ Pose-based Attention draws focus to Hands
			}
	
	\author{Fabien Baradel\\
		Univ Lyon, INSA-Lyon, CNRS, LIRIS\\
		F-69621, Villeurbanne, France\\
		{\tt\small fabien.baradel@liris.cnrs.fr}
		\and
		Christian Wolf\\
		Univ Lyon, INSA-Lyon, CNRS, LIRIS\\
		F-69621, Villeurbanne, France\\
		{\tt\small christian.wolf@liris.cnrs.fr}
		\and
		Julien Mille\\
		Laboratoire d’Informatique de l’Université de Tours (EA 6300), INSA Centre Val de Loire\\
		41034 Blois, France\\
		{\tt\small  julien.mille@insa-cvl.fr}
	}
	
	\maketitle

\begin{abstract}
    \noindent
    We propose a new spatio-temporal attention based mechanism for human action recognition able to automatically attend to the hands most involved into the studied action and detect the most discriminative moments in an action.
    Attention is handled in a recurrent manner employing Recurrent Neural Network (RNN) and is fully-differentiable.
    In contrast to standard soft-attention based mechanisms, our approach does not use the hidden RNN state as input to the attention model.
    Instead, attention distributions are extracted using external information: human articulated pose.
    We performed an extensive ablation study to show the strengths of this approach and we particularly studied the conditioning aspect of the attention mechanism.
We evaluate the method on the largest currently available human action recognition dataset, NTU-RGB+D, and report state-of-the-art results. Other advantages of our model are certain aspects of explanability, as the spatial and temporal attention distributions at test time allow to study and verify on which parts of the input data the method focuses.

	
\end{abstract}

\section{Introduction}

\begin{figure}[t]
    \centering
        \includegraphics[width=8cm]{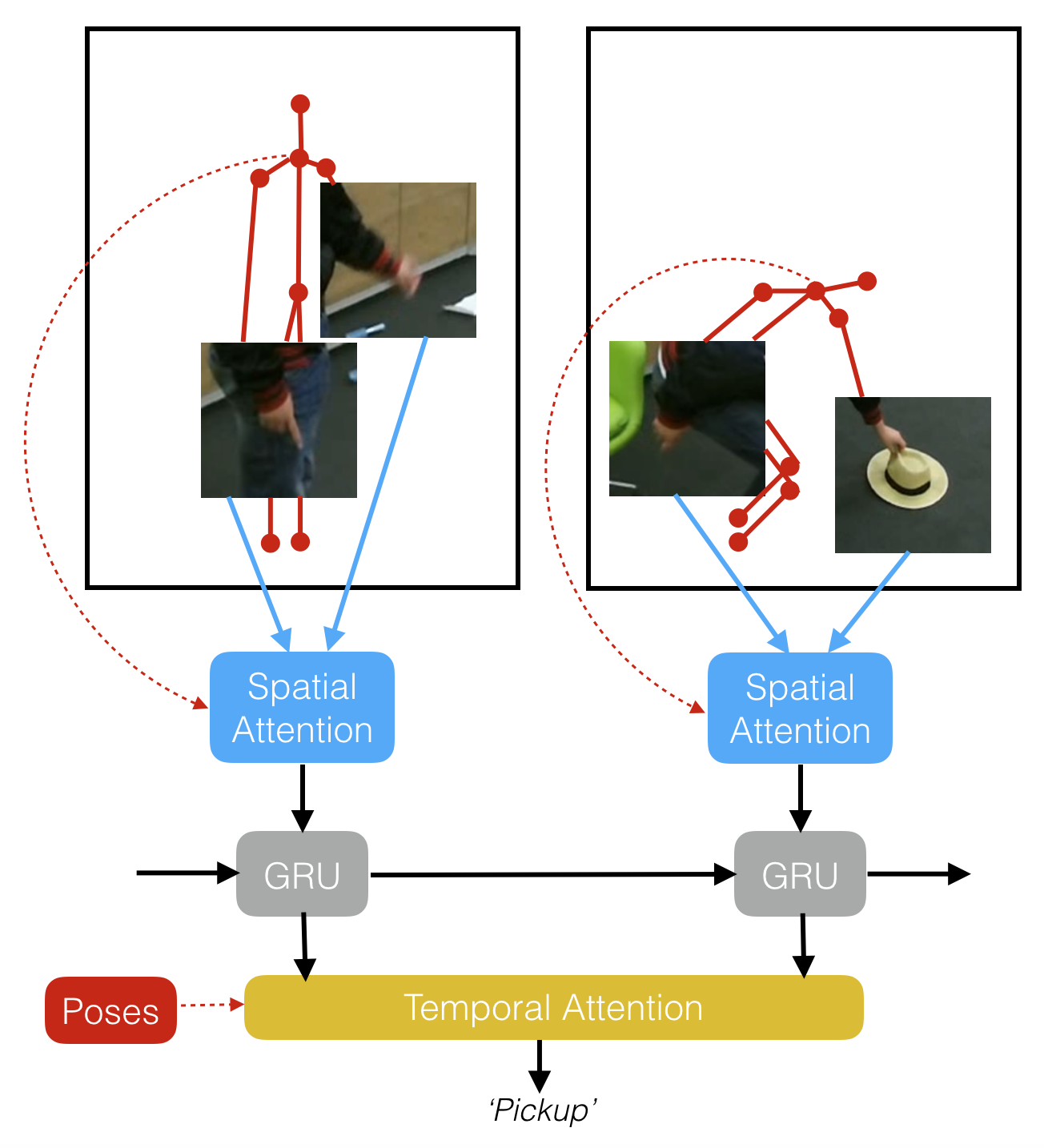}
    \caption{We design a new spatio-temporal mechanism conditioned on pose only able to attend to the most important hands and hidden states.}
    \label{fig:teaser}
\end{figure}

\noindent
Human action recognition is an active field in computer vision with a range of industrial applications, for instance video surveillance, robotics, automated driving and others.
Consumer depth cameras made a huge impact in research and applications since they allow to estimate human articulated poses easily.
Depth input is helpful for solving computer vision problems considered as hard when dealing with RGB inputs only \cite{Han13}.
In this work we address human action recognition in settings where human pose is available in addition to RGB inputs. The RGB stream provides additional rich contextual cues on human activities, for instance on the objects held or interacted with. 

Understanding human behavior remains an unsolved problem compared to other tasks in computer vision and machine learning in general, mainly due to the lack of sufficient data.
Large datasets, such as Imagenet~\cite{Russakovsky2015} for object detection, have allowed powerful deep learning methods to reach super-human performances.
In the field of human action recognition most of the datasets have several hundreds or a few thousand videos.
As a consequence, state-of-the-art approaches on these datasets either use handcrafted features or are suspected to overfit, after years the community spent on tuning methods.
The recent release of large-scale datasets like NTU-RGB-D \cite{Shahroudy2016} (${\sim}$ 57'000 videos) will hopefully lead to better automatically learned representations.


Video understanding is by definition challenging due to its high  dimensional, rich and complex input space.
Most of the time, only a limited area of a video is necessary to get a fine-grained understanding of the occuring action.
Inspired by neuroscience perspectives, models of visual attention \cite{Mnih_NIPS2014,ChoBengioMM2015,Sharma2016a} (see section \ref{sec:relatedworks} for a full discussion) have drawn considerable interest recently.
By attending only to specific areas, parameters are not wasted on input considered as noise for the final task.
We propose a method for human action recognition, which addresses this problem by handling raw RGB input in a novel way.
Instead of taking as input the full RGB frame, we take into account image areas cropped around hands only, whose positions are extracted from full body pose estimated by a middleware.

Our model uses two input streams: (i) an RGB stream called \textit{Spatio-Temporal Attention over Hands (STA-Hands)}, and (ii) a pose stream. 
A key feature of our method is its ability to automatically draw attention to the most important hands at each time step.
Additionally, our approach can also automatically detect the most discriminative hidden RNN states, i.e. most discriminative time instants. 

Beyond giving state-of-the-art results on the NTU dataset, our spatio-temporal mechanism also features certain aspects of explainability. In particular, it gives insights into key choices made by the model at test time in the form of two different attention distributions: a spatial one (which hands are most important at which time instant?) and a temporal one (which time instants are most important?)



The contributions of our work are as follows:
\begin{itemize}


	\item We propose a spatial attention mechanism on human hands on RGB videos which is conditioned on the estimated pose at each time step. 

    \item We propose a temporal attention mechanism which learns how to pool features output from the RNN over time in an adaptive way conditioned on the poses over the full sequence.
    
    \item We show by an extensive ablation study that soft-attention mechanisms (both spatial and temporal) can be done using external variables in contrast to usual approaches which condition the attention mechanism on the hidden RNN state.

\end{itemize}


\section{Related Work}
\label{sec:relatedworks}
\noindent
\myparagraph{Activities, gestures and multimodal data}
Recent gesture/action recognition methods dealing with several modalities typically process 2D+T RGB and/or depth data as 3D.
Sequences of RGB frames are stacked into volumes and fed into convolutional layers at first stages~\cite{Baccouche2011,Ji_PAMI2013, MolchanovYangCVPR2016, NeverovaWolfTaylorNeboutPAMI2016, WuPigouPAMI2016}.
When additional pose data is available, the 3D joint positions are typically fed into a separate network. Preprocessing pose is reported to improve performance in some situations, e.g. augmenting coordinates with velocities and acceleration~\cite{DBLP:conf/iccv/ZanfirLS13}.
Pose normalization (bone lengths and view point normalization) has been reported to help in certain situations \cite{NeverovaWolfTaylorNeboutPAMI2016}.
Fusing pose and raw video modalities is traditionally done as late fusion~\cite{MolchanovYangCVPR2016}, or early through fusion layers~\cite{WuPigouPAMI2016}.
In \cite{LiNeverovaWolfTaylor2017}, fusion strategies are learned together with model parameters by stochastic regularization.







  
\myparagraph{Recurrent architectures for action recognition}
Most recent human action recognition methods are based on recurrent neural networks in some form.
In the variant Long Short-Term Memory (LSTM) ~\cite{Hochreiter1997}, a gating mechanism over an internal memory cell learns long-term and short-term dependencies in the sequential input data.
Part-aware LSTMs ~\cite{Shahroudy2016} separate the memory cell into part-based sub-cells and let the network learn long-term representations individually for each part, fusing the parts for output. 
Similarly, Du {\it et al}~\cite{Du_CVPR2015} use bi-directional LSTM layers which fit anatomical hierarchy.
Skeletons are split into anatomically-relevant parts (legs, arms, torso, {\it etc}), so that each subnetwork in the first layers gets specialized on one part. Features are progressively merged as they pass through layers. 

Multi-dimensional LSTMs \cite{Graves2DLSTM2009} are models with multiple recurrences from different dimensions.
Originally introduced for images, they also have been applied to activity recognition from pose sequences~\cite{Liu2016}.
The first dimension is time, while the second one is a topological traversal of the joints in a bidirectional depth-first search, which preserves the neighborhood relationships in the graph.




\myparagraph{Attention mechanisms}
Human perception focuses selectively on parts of the scene to acquire information at specific places and times.
In machine learning, this kind of process is referred to as attention mechanism, and has drawn increasing interest when dealing with languages, images and other data.
Integrating attention can potentially lead to improved overall accuracy, as the system can focus on parts of the data, which are most relevant to the task.
 
In computer vision, visual attention mechanisms date as far back as the work of Itti {\it et al} for object detection~\cite{Itti_PAMI1998} and has been inspired by works from the neuroscience community \cite{Jonides1983}. 
Early models were highly related to saliency maps, i.e. pixelwise weighting of image parts that locally stand out. No learning was involved.
Larochelle and Hinton~\cite{Larochelle_NIPS2010} pioneered the incorporation of attention into a learning architecture by coupling Restricted Boltzmann Machines with a foveal representation.

More recently, attention mechanisms were gradually categorized into two classes.
\emph{Hard attention} takes hard decisions when choosing parts of the input data.
This leads to stochastic algorithms, which cannot be easily learned through gradient descent and back-propagation.
In a seminal paper, Mnih {\it et al}~\cite{Mnih_NIPS2014} proposed visual hard-attention for image classification built around a recurrent network, which implements the policy of a virtual agent.
A reinforcement learning problem is thus solved during learning~\cite{Williams1992}.
The model selects the next location to focus on, based on past information.
Ba et al ~\cite{Ba-attention-2015} improved the approach to tackle multiple object recognition.
In \cite{Kuen_CVPR2015}, a hard attention model generates saliency maps. Yeung {\it et al}~\cite{Yeung_CVPR2016} use hard-attention for action detection with a model, which decides both which frame to observe next as well as when to emit an action prediction. 

On the other hand, \emph{soft attention} takes the entire input into account, weighting each part of the observations dynamically.
The objective function is usually differentiable, making gradient-based optimization possible. Soft attention was used for various applications such as neural machine translation~\cite{Bahdanau_ICLR2015, Kim_ICLR2017} or image captioning~\cite{Xu_ICML2015}.
Recently, soft attention was proposed for image \cite{ChoBengioMM2015} and video understanding ~\cite{Sharma2016a,Song2016,yeung2015every}, with spatial, temporal and spatio-temporal variants.
Sharma {\it et al}~\cite{Sharma2016a} proposed a recurrent mechanism for action recognition from RGB data, which integrates convolutional features from different parts of a space-time volume.
Yeung et al. report a temporal recurrent attention model for dense labeling of videos~\cite{yeung2015every}.
At each time step, multiple input frames are integrated and soft predictions are generated for multiple frames.
An extended version of this work has been proposed \cite{Li16} by also taking into account the optical flow.
Bazzani {\it et al} \cite{Bazzani_ICLR2017} learn spatial saliency maps represented by mixtures of Gaussians, whose parameters are included into the internal state of a LSTM network. 
Saliency maps are then used to smoothly select areas with relevant human motion.
Song {\it et al} \cite{Song2016} propose separate spatial and temporal attention networks for action recognition from pose. At each frame, the spatial attention model gives more importance to the joints most relevant to the current action, whereas the temporal model selects frames.


Up to our knowledge, no attention model has yet taken advantage of articulated pose for attention over RGB sequences. 

Our method has slight similarities with ~\cite{Mnih_NIPS2014} in that crops are done on locations in each frame.
However, in our case, these operations are not learned, they depend on pose.
On the other hand, we learn a soft-attention mechanism, which dynamically weights features from several locations.
The mechanism is conditioned on pose, which allows it to steer its focus depending on motion.


\section{Proposed Model}
\label{sec:proposed-model}

\noindent
A single or multi-person action is described by a sequence of two modalities: the set of RGB input images $\bI{=}\{ I_t \}$, and the set of articulated human poses $\bx{=}\{ \bx_t \}$.
Both signals are indexed by time $t$.
Poses $\bx_t$ are defined by 3D coordinates of joints.
We propose a hands spatio-temporal attention based mechanism conditioned on pose.
This stream processes RGB data $\bI$ and also uses pose information $\bx$ (human body joint locations and their dynamics).
Our two-stream model comprises the aggregation of the streams presented below.

\subsection{SA-Hands: Spatial Attention on Hands}
\noindent
Most of the existing approaches for human action recognition focus on pose data,
which provides good high level information of the body motion in an action but somewhat limits feature extraction.
A large number of actions such as \emph{Reading}, \emph{Writing}, \emph{Eating}, \emph{Drinking} share the same body motion and can be differentiated only by looking at manipulated objects and hands shapes.
Performing fine-grained understanding of human actions can be handled by extracting cues from the RGB streams.

To solve this, we define a glimpse sensor able to crop images around hands at each time step.
This is motivated by the fact that humans perform most of their actions using their hands.
The cropping operation is done using the pixel coordinates of each hand detected by the middleware (up to 4 hands for human interactions between 2 people).
The glimpse operation is fully-differentiable since the exact locations are inputs to the model.
The goal is to extract information about hand shapes and about manipulated objects and to draw attention to specific hands.

\begin{figure}[t!] \centering
        \includegraphics[width=8cm]{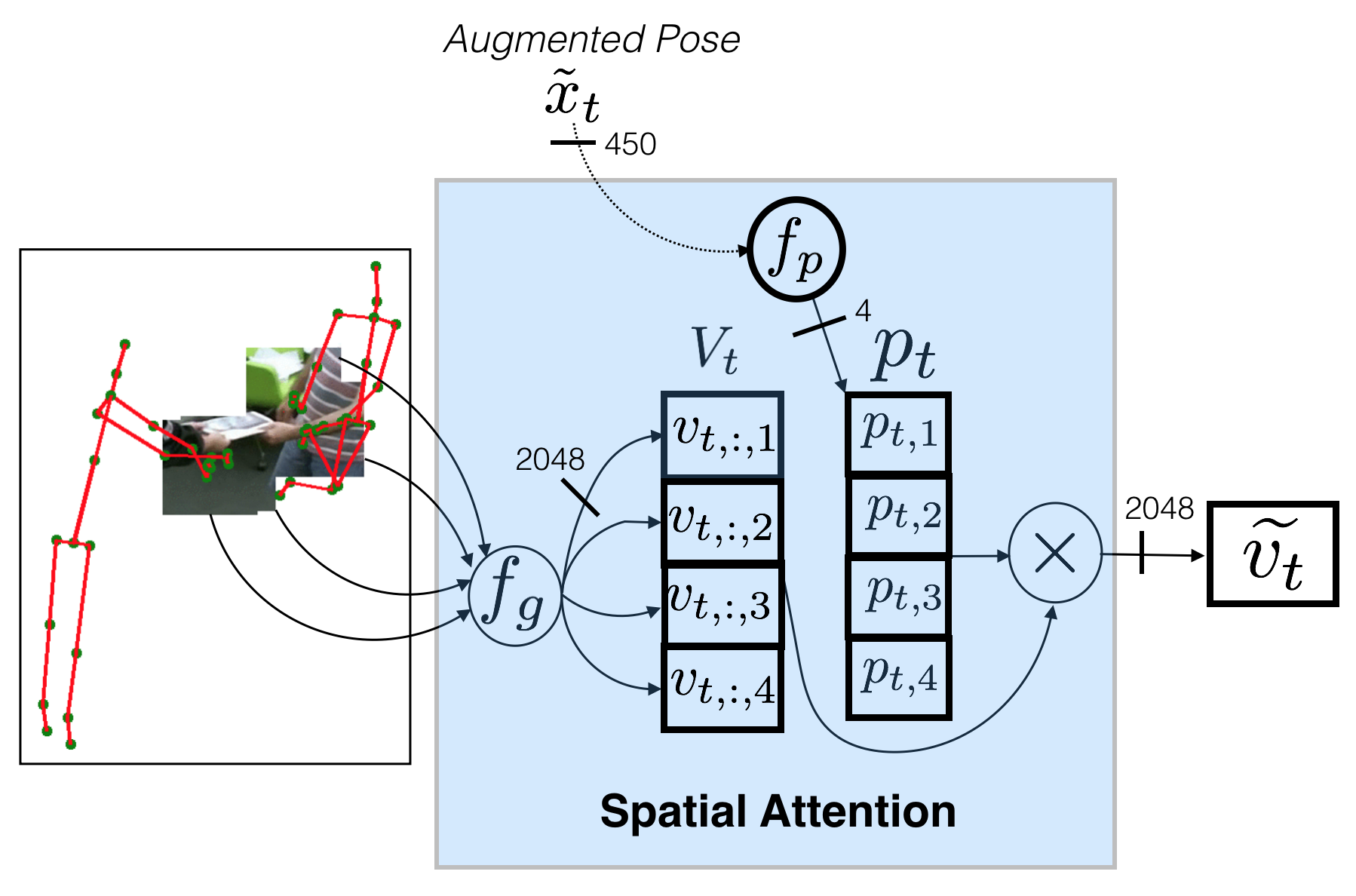}
        \caption{\label{fig:sattention} The spatial attention mechanism: \textit{SA-Hands}.}
\end{figure}

The glimpse representation for a given hand $i$ is a convolutional network $f_g$ with parameters $\theta_g$ (e.g. a pretrained Inception v3), taking as input a crop taken from image $I_t$ at the position of hand $i$:
\begin{equation}
\bv_{t,:,i} = f_g (\textrm{crop}(I_t,\textrm{hand}_i);\theta_g)  {\small \quad \quad i{=}\{1,\dots 4\} }
\end{equation}
Here and in the rest of the paper, subscripts of mappings $f$ and their parameters $\theta$ choose a specific mapping, they are not indices.
Subscripts of variables and tensors are indices.
$\bv_{t,:,i}$ is a (column) feature vector for time $t$ and hand $i$.
For a given time $t$, we stack the vectors into a matrix $\bV_t {=}\{ \bv_{t,:,i} \}$, where $i$ is the index over hand joints and $j$ the index over the feature dimensions .
$\bV_t$ is a matrix (a 2D tensor), since $t$ is fixed for a given instant.
 
A recurrent model receives inputs from the glimpse sensor sequentially and models the information from the seen sequence with a componential hidden state $\bh_t$:
\begin{equation}
\bh_t = f_h (\bh_{t-1}, \tilde{\bv}_t; \theta_h)
\label{eq:lstm}
\end{equation}
We select the GRU as our recurrent function $f_h$.
To keep the notation simple, we omitted the gates from the equations.
The input fed to the recurrent network is the context vector $\tilde{\bv}_t$, defined further below, which corresponds to an integration of the different features vectors extracted from hands in $\bV_t$. 

An obvious choice of integration are simple functions like sums and concatenations.
While the former tends to squash feature dynamics by pooling strong feature activations in one hand with average or low activations in other hands, the latter leads to high capacity models with low generalization.

We employ a soft-attention mechanism which dynamically weighs the integration process through a distribution $\bp_t$, determining how much attention hand $i$ needs with a calculated weight $\bp_{t,i}$.
We define the \textit{augmented pose} vector $\tilde{\bx}_t$ defined by the concatenation of the current pose $\bx_t$, the acceleration $\dot{\bx_t}$ and the velocity $\ddot{\bx}_t$ for each joint over time.
At each time step, $\tilde{\bx}_t$ gives a brief overview of human poses on the scene and their dynamics.
In contrast to existing soft-attention based mechanisms \cite{Sharma2016a,Xu2015,Li16}, our attention distribution does not depend on the previous hidden state $\bh_{t-1}$ of the recurrent network, but exclusively on an external information defined above: the \textit{augmented pose} $\tilde{\bx}_t$.

Finally, spatial attention weights $\bp_t$ are given through a learned mapping with parameters $\theta_p$:
\begin{equation}
\bp_t = f_p (\tilde{\bx}_t; \theta_p)
\label{eq:sattention}
\end{equation}
Remark that if we replace $\tilde{\bx}_t$ by $\bh_{t-1}$ in equation \ref{eq:sattention} we get the usual soft-attention mechanism by conditioning the attention weights on the hidden state \cite{Sharma2016a}.
Attention distribution $\bp_t$ and features $\bV_t$ are integrated through a linear combination as
\begin{equation}
\tilde{\bv}_t = \bV_{t} \bp_{t} \ ,
\end{equation}
which is input to the GRU network at time $t$ (see eq. (\ref{eq:lstm})).
The conditioning on the \textit{augmented pose} in \ref{eq:sattention} is important, as it provides valuable body motion information at each timestep (see the ablation study in the experimental section).

We refer to this model as \textit{SA-Hands} in our table.
For a better understanding of this module, a visualization can be found in Figure \ref{fig:sattention}.

\begin{figure}[t!] \centering
        \includegraphics[width=6cm]{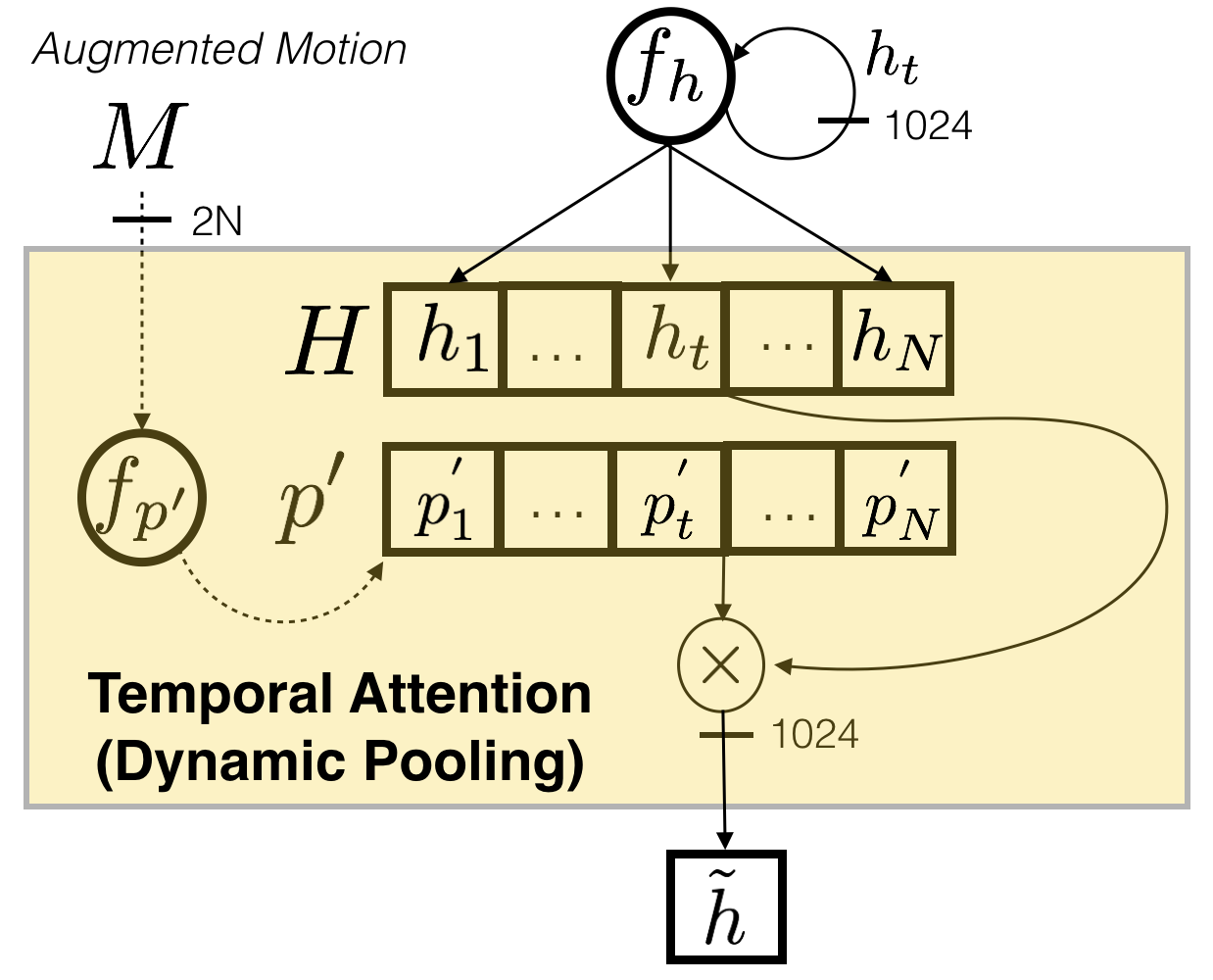}
        \caption{\label{fig:tattention} The temporal attention mechanism:  \textit{TA-Hands}}.
\end{figure}

\begin{figure}[t!] \centering
	\includegraphics[width=8cm]{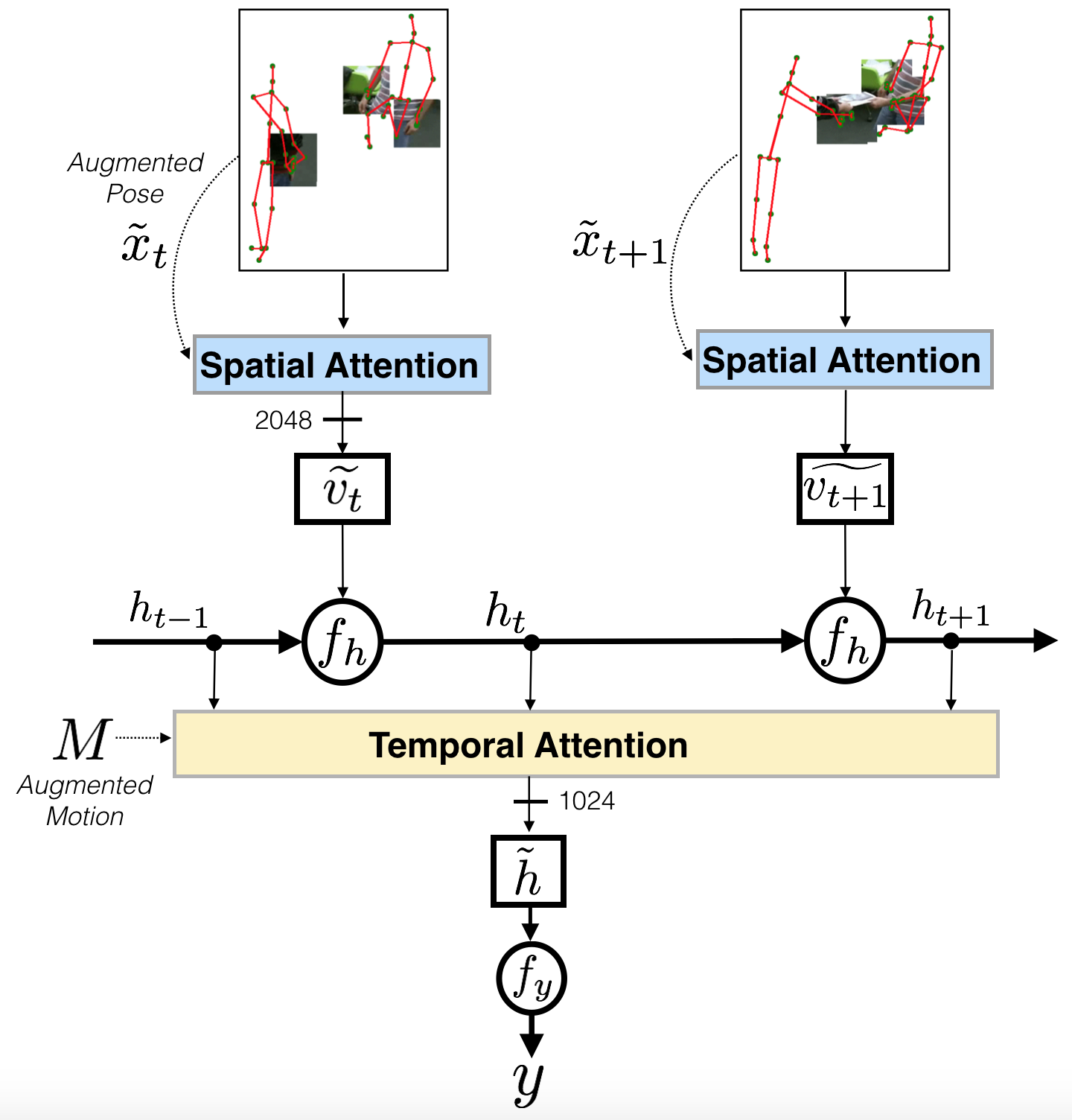}
	\caption{\label{fig:fullattention} The spatio-temporal attention mechanism: \textit{STA-Hands}. The spatial mechanism is detailed in figure \ref{fig:sattention} and the temporal one is detailed in figure \ref{fig:tattention}}
\end{figure}

\begin{figure*}[t] \centering
    \includegraphics[width=12cm]{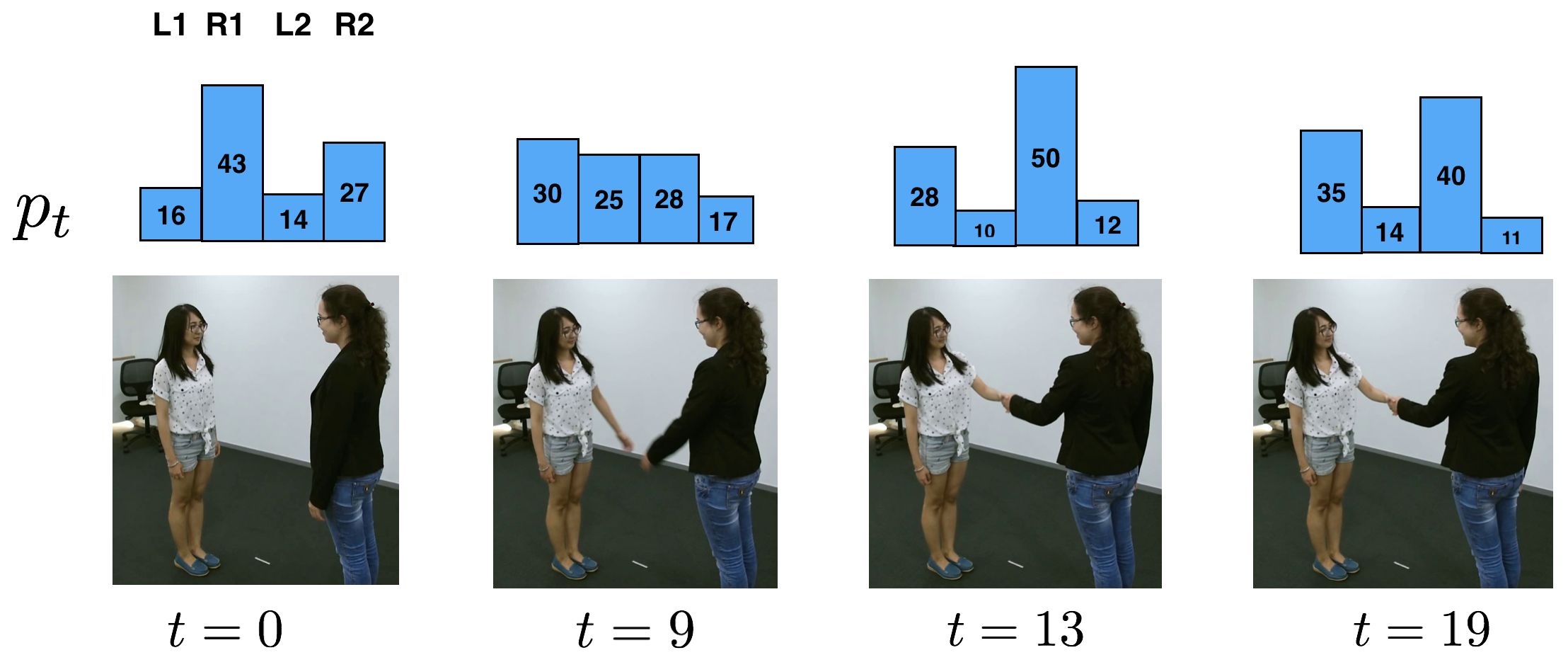}
    \caption{\label{fig:attentionexample} Spatial attention over time: shaking hands will make the attention shift to hands in action.}
\end{figure*}

\subsection{TA-Hands: Temporal Attention on Hidden States}

\noindent
Recurrent models can provide predictions for each time step $t$ by performing a mapping directly from the hidden state $\bh_t$.
Some hidden states are more discriminative than other ones.
Following this idea we perform a temporal pooling on the hidden state level in an adaptive way.
At the end of the sequence an attention mechanism automatically gives weights for each hidden state.

The hidden states for all instants $t$ of the sequence are stacked into a 2D matrix $\bH{=}\{\bh_{j,t}\}$, where $j$ is the index over the hidden state dimension.
A temporal attention distribution $\bp'$ is predicted through a learned mapping to automatically identify the most important hidden states (i.e. the most important time instants $t$).
To be efficient, this mapping should have seen the full sequence before giving a prediction for an instant $t$, as giving a low weight to features at the beginning of a sequence might be caused by the need to give higher weights to features at the end.

To keep the model simple, we benefit from the fact that sequences are of fixed length.
We define a statistic called \textit{augmented motion} $\bm_t$ given by the sum of the absolute acceleration and the sum of the absolute velocity of all body joints at each time step $t$.
$\bm_t$ is a vector of size 2 
and we obtain $M$ by stacking all $\bm_t$.
$M$ gives a good overview of when most important moments occur.
Our assumption is that higher values of $\bm_t$ indicate more useful instants 
$t$. But of course the network can learn more complex mappings reacting to more complex motions or poses.
The temporal attention weights are given by the mapping:
\begin{equation}
\bp' = f'_p (\bM; \theta'_p)
\end{equation}
This attention is used as weight for adaptive temporal pooling of the features $\bH$, i.e. 
$$
\tilde{\bh} = \bH \bp' \quad .
$$
We called this module \textit{TA-Hands}.
A visualization of the module can be found in figure \ref{fig:tattention}.

The spatial and temporal attention mechanism are independent of each other.
When both are combined we call the model \textit{Spatio-Temporal Attention over Hands (STA-Hands)}.
A visualization of the overall RGB stream can be found in figure \ref{fig:fullattention}.

\begin{figure*}[t] \centering
	\includegraphics[width=12cm]{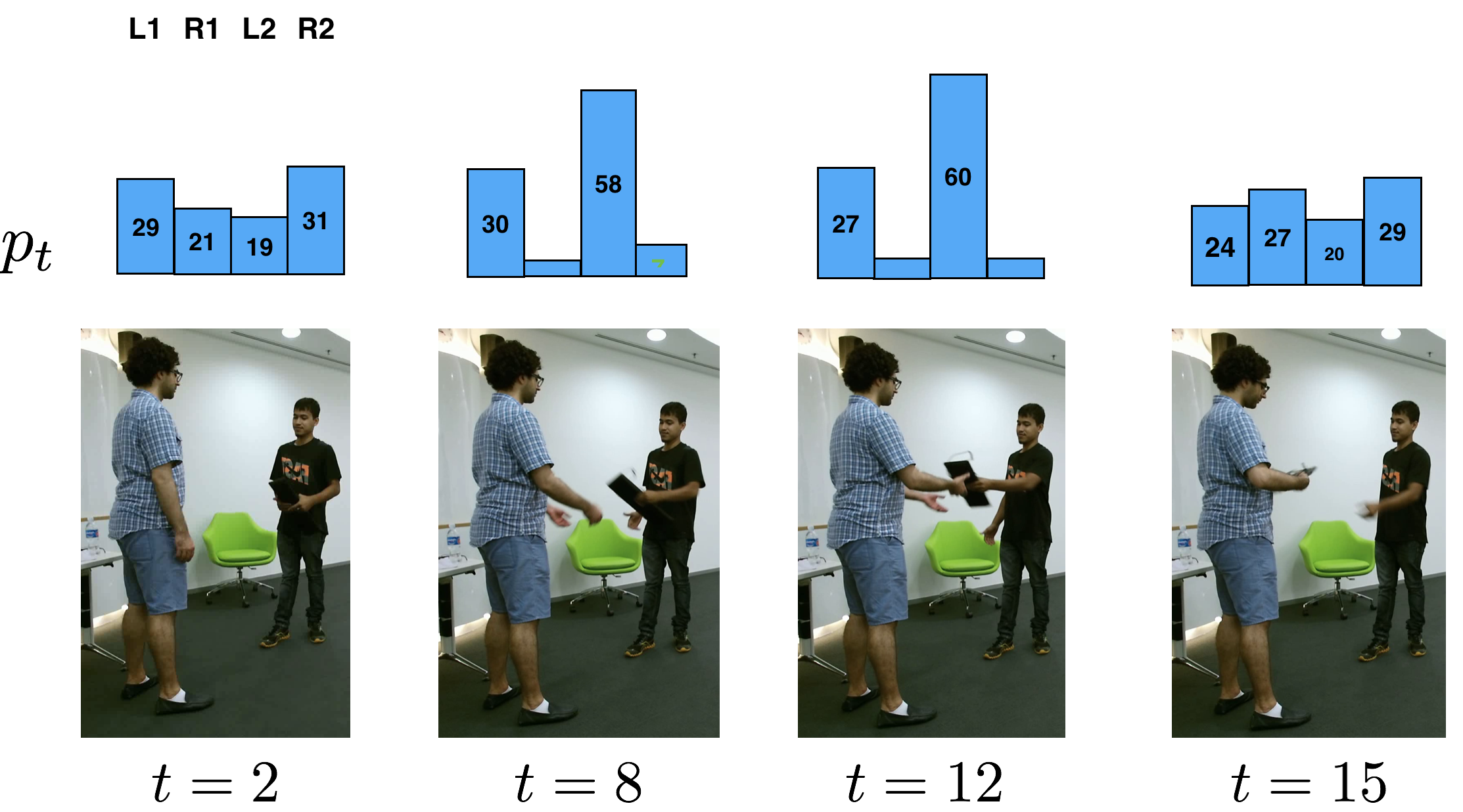}
	\caption{\label{fig:attentionexample2} Spatial attention over time: giving something to other person will make the attention shift to the active hands in the action.}
\end{figure*}

\emph{Related work} --- note that most current approaches in sequence classification proceed by temporal pooling of individual predictions, e.g. through a sum or average \cite{Sharma2016a} or even by taking predictions of the last time step.  We show that it can be important to perform this pooling in an adaptive way.
In recent work on dense activity labeling, temporal attention for dynamical pooling of LSTM logits has been proposed \cite{yeung2015every}. In the context of sequence-to-sequence alignment, temporal pooling has been addressed with bi-directional recurrent networks \cite{Bahdanau2014}.

\subsection{Deep GRU: Gated Recurrent Unit on Poses}
\label{subsec:gru_pose}

\noindent
Above, the pose information was used as valuable input to the RGB stream. Articulated pose is also used directly for classification in a second stream, the pose stream. We process the sequence of pose, where at each time step $t$, $ \bx_t $ is a vector which represents the concatenation of 3D coordinates of joints of all subjects. The raw pose vectors are input into a RNN.

In particular, we learn a pose network $f_{sk}$ with parameters $\theta_{sk}$ on this input sequence $\bx$, resulting in a set of hidden state representation $\bh^{sk}{=}\{ \bh_t^{sk} \}$:
\begin{equation}
\bh^{sk}_t = f_{sk} (\bh^{sk}_{t-1}, \bx_t; \theta_{sk})
\end{equation}
We call this baseline on poses \textit{Deep GRU} in our tables.

\subsection{Stream fusion}

\noindent
Each stream, pose and RGB, leads to its own features, respectively $\bh^{sk}$ for the pose stream and $\tilde{\bh}$ for the RGB stream.
Each representation is classified with its own set of parameters using a standard classification approach as defined further below in \ref{sec:train}.
We fuse both streams on logit level by summing.
More sophisticated techniques, such as features concatenation and learned fusion \cite{NeverovaWolfTaylorNeboutPAMI2016} have been evaluated and rejected.

\section{Network architectures and Training}
\label{sec:archtrain}

\noindent
\myparagraph{Architectures}
The pose network $f_{sk}$ consists of a stack of 3 GRU each with an hidden state of size 150.

The glimpse sensor $f_g$ is implemented as an Inception V3 network \cite{Szegedy2016}.
Each vector $\bv_{t,:,i}$ corresponds to the last layer before output and is of size 2048.
The GRU network $f_h$ has a single recurrent layer with 1024 units.
The spatial attention network $f_p$ is an MLP with a single hidden layer of 256 units with ReLu activation.
The temporal attention network $f'_p$ is an MLP with a single hidden layer of 32 units with ReLu activation.
Output layers of attention networks $f_p$ and $f'_p$ use the softmax activation in order to get the sum of the attention weights equal to 1.
The full model (without glimpse sensor $f_g$) has 10 millions trainable parameters.

\myparagraph{Training}
\label{sec:train}
All classification are done using a simple fully-connected layer followed by a softmax activation and trained with cross-entropy loss.
For the pose stream \textit{Deep GRU} the classification is learned from all the hidden states $\bh^{sk}_t$.
At test time we average the predictions given by each time step since it gives better results than taking predictions from the last hidden state.

For the RGB stream, classification using \textit{STA-Hands} is learned from the feature vector $\tilde{\bh}$.
When the temporal attention (i.e.\textit{TA-Hands}) is not employed in the RGB stream we follow the same settings as described for the pose stream.
The glimpse sensor $f_g$ is pretrained on the ILSVRC 2012 data~\cite{Russakovsky2015} and is frozen during training.
Both spatial $\bp$ and temporal attention weights $\bp{'}$ are initialized to be equal for each input modality.
This setup leads to faster convergence and better stability during training.

\section{Experiments}
\label{sec:experiments}

\noindent
The proposed method has been evaluated on the largest human action recognition dataset: NTU RGB+D.
We extensively tested all aspects of our model by conducting an ablation study.
This leads to a proper understanding of the choice of our proposed new spatio-temporal mechanism and specially its conditioning aspect.

The NTU RGB+D Dataset (NTU) ~\cite{Shahroudy2016} has been acquired with a Kinect v2 sensor and contains more than 56K videos and 4 millions frames with 60 different activities including individual activities, interactions between 2 people and health related events. The actions have been performed by 40 subjects and with 80 viewpoints.
The 3D coordinates of 25 body joints are provided in this dataset.
We follow the cross-subject and cross-view split protocol from~\cite{Shahroudy2016}.
Due to the large amount of videos, this dataset is highly suitable for deep learning modeling.

\begin{table}
	\begin{center}
		\begin{tabular}{cccccc}
			\arrayrulecolor{cwblue1} \toprule
			Methods                                  & {\footnotesize Pose} &  {\footnotesize RGB} & CS & CV & Avg \\ 
            \arrayrulecolor{cwblue1} \toprule
			Lie Group \cite{Vemulapalli_2014_CVPR}     & X & -& 50.1          & 52.8       & 51.5    \\ 
			Skeleton Quads \cite{Evangelidis-ICPR-2014} & X & -&38.6          & 41.4       & 40.0    \\ 
			Dynamic Skeletons \cite{Hu_CVPR2015}      & X & -&60.2          & 65.2       & 62.7    \\ 
			HBRNN \cite{Du_CVPR2015}     & X & -&59.1          & 64.0       & 61.6    \\ 
			Deep LSTM \cite{Shahroudy2016}        & X & -&60.7          & 67.3       & 64.0    \\ 
			Part-aware LSTM \cite{Shahroudy2016}  & X & -&62.9          & 70.3       & 66.6    \\ 
			ST-LSTM + TrustG.  \cite{Liu2016}    & X & -&69.2          & 77.7       & 73.5    \\ 
			STA-LSTM   \cite{Song2016}         & X & -&73.2          & 81.2       & 77.2    \\ 
			GCA-LSTM   \cite{Liu_2017_CVPR}         & X & -&74.4         & 82.8       & 78.6    \\ 
			JTM   \cite{Wang}          & X & -& 76.3   & 81.1   &  78.7 \\ \hline
			MTLN \cite{Ke_2017_CVPR} & X & - & 79.6 & 84.8 & 82.2 \\ \hline
            DSSCA - SSLM \cite{Shahroudy20162} & X & X & 74.9 & - & -  \\ \hline
			\textbf{Deep GRU [A]}    & X & -& \textbf{68.0}        & \textbf{74.2}     & \textbf{71.1}   \\ 
			\textbf{STA-Hands [B]}    & $\circ$ & X & \textbf{73.5}        & \textbf{80.2}     & \textbf{76.9}   \\ 
            \textbf{A+B}    & X & X& \textbf{82.5}       &   \textbf{88.6} & \textbf{85.6}   \\
            \arrayrulecolor{cwblue1} \bottomrule
		\end{tabular}
	\end{center}
	\caption{Results on the NTU RGB+D dataset with Cross-Subject (CS) and Cross-View (CV) settings (accuracies in \%, $\circ$ means that pose is only used for the attention mechanism).}
	\label{NTU}
\end{table}

%

\begin{table*}[t]
    \begin{center}
        \begin{tabular}{ccccccc}
            \arrayrulecolor{cwblue1} \toprule
            Methods & \multicolumn{2}{c}{Spatial Attention} & Temporal Attention & CS& CV & Avg \\ 
             & {\footnotesize Hidden state} & {\footnotesize Augmented Pose} & {\footnotesize Augmented Pose}&&& \\
            \arrayrulecolor{cwblue1} \toprule
            Sum                                  &               -            &        -          &        - & 68.3      &   74.6     &     71.5 \\
            Concat                              &              -             &       -           &        - & 68.9      &     75.2  &     72.0 \\ \hline
            \multirow{3}{*}{SA-Hands}         & X                         &      -           &        -  & 69.8      &   76.2     &    73.0 \\
            &         -                  & X                &        - & \textbf{71.0}      &    \textbf{78.9}    &  \textbf{75.0}  \\
            & X                         & X                &      - & 70.5      &     76.6        &  73.6\\ \hline
            TA-Hands     &- & -       &        X&\textbf{71.1}      &  \textbf{78.5}     &   \textbf{74.8} \\
            \arrayrulecolor{cwblue1} \toprule
            \multirow{3}{*}{STA-Hands} & X                         &        -           &        X& 72.2      &   77.8     &     75.0\\
            &        -                   & X                 &        X& \textbf{73.5}      &   \textbf{80.2}     &  \textbf{76.9}\\
            & X                         & X                 &        X& 72.8      &     78.3 & 75.6 \\
            \arrayrulecolor{cwblue1} \bottomrule 
        \end{tabular}
    \end{center}
    \caption{Effects of the conditioning on the spatial attention and the temporal attention (RGB stream only, accuracies in \%).}
    \label{attention_rgb}
\end{table*}


\begin{table*}[t]
	\begin{center}
		\begin{tabular}{c|cccccc}
			\arrayrulecolor{cwblue1} \toprule
			RGB stream methods & 
			\multicolumn{2}{c}{Spatial Attention} & Temporal Attention & CS& CV & Avg \\
			&   {\footnotesize Hidden state} & {\footnotesize Augmented Pose}  & {\footnotesize Augmented Motion} &&& \\
			\arrayrulecolor{cwblue1} \toprule
			Sum-Hands& - &  -&-&  79.5& 85.9  &  82.8     \\ 
			\arrayrulecolor{cwblue1} \toprule
			\multirow{3}{*}{SA-Hands} &X&- &-&80.5 & 86.8 & 83.7\\ 
			&-&X&-&81.4 & 87.4& 84.4  \\ 
			& X & X &-& 81.0 & 86.9 &84.0     \\ 
			\arrayrulecolor{cwblue1} \bottomrule 
			TA-Hands& - & - &X& 80.8 & 87.6 & 84.2     \\ 
			\arrayrulecolor{cwblue1} \bottomrule 
			\multirow{3}{*}{STA-Hands} &X&- &X& 81.4 & 87.4 & 84.4\\ 
			&-&X&X&\textbf{82.5} & \textbf{88.6}&   \textbf{85.6}  \\ 
			& X & X &X& 81.6 & 88.0 & 84.8     \\ 
			\arrayrulecolor{cwblue1} \bottomrule 
		\end{tabular}
	\end{center}
	\caption{Effects of conditioning the spatio-temporal attention on different latent variables in the RGB stream for the two-stream model (accuracies in \% on NTU). The pose stream is always the same: (\textit{Deep GRU}) for every row.}
	\label{attention_two_stream}
\end{table*}

\myparagraph{Implementation details}
Following \cite{Shahroudy2016}, we cut videos into sub sequences of 20 frames and sample sub-sequences.
During training a single sub-sequence is sampled. During testing 5 sub-sequences are extracted and logits are averaged.
We apply a normalization step on the joint coordinates by translating them to a body centered coordinate system with the "middle of the spine'' joint as the origin.
If only one subject is present in a frame, we set the coordinates of the second subject to zero.
We crop sub images of static size $50{\times}50$ on the positions of the hand joints (pixel locations of each hands are given by the middleware).
Cropped images are then resized to $299{\times}299$ and fed into the Inception model. 

Training is done using the Adam Optimizer \cite{AdamOptimization2015} with an initial learning rate of 0.0001.
We use minibatches of size 32, dropout with a probability of 0.5 and train our model up to 100 epochs.
Following \cite{Shahroudy2016}, we sample 5\% of the initial training set as a validation set, which is used for hyper-parameter optimization and for early stopping.
All hyperparameters have been optimized on the validation sets.


\myparagraph{Comparisons to the state-of-the-art}
We show comparisons of our model to the state-of-the-art methods in table \ref{NTU}.
 We achieve state of the art performance on the NTU dataset with the two-stream model even if we intentionally implemented a weak model, \textit{Deep GRU}, on the pose stream.
 That shows the strength of our RGB stream called \textit{STA-Hands} at extracting cues.
Comparing one by one our two streams (RGB vs pose) demonstrates that \textit{STA-Hands} gets better results than \textit{Deep GRU}.
 
 We have to keep in mind that the pose is used as external data in our RGB stream but only for the cropping operation around hands and for computing the attention distributions.
 Poses are never directly fed as input to the GRU in \textit{STA-Hands} for updating the hidden state.
The purpose of \textit{STA-Hands} is to extract cues from hand shapes or manipulated objects.
By its design choice \textit{STA-Hands} is not able to extract body motion since pose is only used for computing an attention distribution over hands.
 However this stream achieves better performance than the pose one, which shows that RGB data should not be put aside for human action recognition.


We conducted extensive ablation studies to understand the impact of our design choices on the full model, and in particular on the spatial attention mechanism \textit{STA-Hands}.

%

\myparagraph{Conditioning the spatial attention}
\noindent
Conditioning the spatial attention on the statistics of the pose (\textit{augmented pose}) at each time step is a key design choice, as shown in table \ref{attention_rgb} (\textit{SA-Hands} rows).
Compared to usual soft-attention mechanisms, which condition attention on the hidden state, we gain 2 points on average (75.0 vs 73.0).
Interestingly, conditioning using both the hidden state and the pose statistics deteriorates the performances (75.0 vs 73.6) showing that different kinds of information are contained in these two latent variables.
The recurrent unit is not able to combine those two cues or at least ignore the hidden state.
We can conclude that the \textit{augmented pose} is a better latent variable for weighting the spatial attention compared to the internal hidden state of the GRU.
Compared to simple baselines like summing the different inputs, our methods improve the average accuracy by 3.5 points (75.0 vs 71.5).
This opens new perspectives for creating attention mechanisms conditioned on new latent variables which can be external to the GRU (but highly correlated to the inputs and to the final task).

\myparagraph{Effect of the temporal attention}
\noindent
Weighted integration of the hidden states over time seems to be an important design choice, as shown in table \ref{attention_rgb}.
Compared to classical baselines, like averaging the predictions, we improve performance by 3.3 points in average (74.8 vs 71.0).
Taking only the final predictions even leads to worst performance.
Again we can see that pose and its statistics, in this case the \textit{augmented motion}, are good latent variables for computing the temporal attention weights, although they are external to the input data but highly correlated.

\myparagraph{A powerful spatio-temporal attention mechanism}
\noindent
We show consistent results by combining spatial and temporal attention trained end-to-end.
Conditioning the spatial and temporal attention mechanisms on statistics of the pose (respectively \textit{augmented pose} and \textit{augmented motion}) leads to the best results.
In average we gain up to 5.4 and 4.9 points compared to the baseline without any attention modules like summing or concatenating the inputs (76.9 vs 71.5 and 72.0).

\myparagraph{Impact of the attention on the two stream model}
Again we get consistent results when going from RGB stream only to two-stream model (pose and RGB streams).
Even if both streams are trained separately and fused at the logit level they extract complementary features.
Spatial attention seems to be more important than temporal one (85.6 vs 84.2).
Compared to baseline like summing inputs on the RGB stream, our full spatio-attention mechanism conditioned on poses beats the baseline by 2.8 points on the two-stream model.

\myparagraph{Runtime}
For a sequence of 20 frames, we get the following runtimes for a single Titan-X (Maxwell) GPU and an i7-5930 CPU: A full prediction from Inception features takes $1.4$ms including pose feature extraction. This does not include RGB pre-processing, which takes additional 1sec (loading Full-HD video, cropping sub-windows and extracting Inception features). Classification can thus be done close to real-time. Fully training one model (w/o Inception) takes $\sim$4h on a Titan-X GPU. Hyper-parameters have been optimized on a computing cluster with 12 Titan-X GPUs.
The proposed model has been implemented in Tensorflow.

\myparagraph{Pose noise}
Crops are performed on hand locations given by the middleware. 
In case of noise, crops could end up not being on hands. We saw, that the attention model can cope with this problem in many cases.

\section{Conclusion}
\label{sec:conclusion}

\noindent
We propose a new method for dealing with RGB video data for human action recognition given pose.
A soft-attention mechanism based on crops around hand joints allows the model to collect relevant features on hand shapes and on manipulated objects from more relevant hands.
Adaptive temporal pooling further increases performance.
We show that conditioning attention mechanisms on pose leads to better results compared to standard approaches which conditioned on the hidden state.
Our method on RGB stream can be seen as a plugin which can be added to any powerful pose stream.
Our two-stream approach shows state-of-the-art results on the largest human action recognition even by employing a weak pose stream.

\section{Acknowledgements}
\noindent
This work was funded under grant ANR Deepvision (ANR-15-CE23-0029), a joint French/Canadian call by ANR and NSERC.

{\small
\bibliographystyle{ieee}
\bibliography{hands_on_action}
}

\end{document}